# Automatic Teaching Platform on Vision Language Retrieval Augmented Generation

Ruslan Gokhman, Jialu Li, Youshan Zhang
Artificial Intelligence
Graduate Computer Science and Engineering Department, Yeshiva University, NY, USA
rgokhman@mail.yu.edu, jli10@mail.yu.edu, youshan.zhang@yu.edu

*Abstract* - **Automating teaching presents unique challenges, as replicating human interaction and adaptability is complex. Automated systems cannot often provide nuanced, real-time feedback that aligns with students' individual learning paces or comprehension levels, which can hinder effective support for diverse needs. This is especially challenging in fields where abstract concepts require adaptive explanations. In this paper, we propose a vision language retrieval augmented generation (named VL-RAG) system that has the potential to bridge this gap by delivering contextually relevant, visually enriched responses that can enhance comprehension. By leveraging a database of tailored answers and images, the VL-RAG system can dynamically retrieve information aligned with specific questions, creating a more interactive and engaging experience that fosters deeper understanding and active student participation. It allows students to explore concepts visually and verbally, promoting deeper understanding and reducing the need for constant human oversight while maintaining flexibility to expand across different subjects and course material.**

## 1. Introduction

In the evolving landscape of digital education, the integration of advanced technologies into learning environments represents a transformative shift aimed at enriching educational delivery and engagement. While interactive learning platforms have broadly embraced technological innovations, they often underutilize the potential of visual content - a critical element that enhances understanding and engagement through intuitive and dynamic presentations.

However, current educational technologies often fall short of fully exploiting the capabilities of visual content, which is a rich medium for conveying complex information intuitively and engagingly. Traditional learning management systems and e-learning tools focus primarily on text-based interactions, which can hinder the learning process for visual learners and reduce engagement in inherently visual subjects, such as science, engineering, and the arts. Visual Question Answering (VQA) systems emerge as a groundbreaking solution to these challenges. By combining computer vision and natural language processing, VQA systems enable the analysis of visual data and the generation of natural language answers to questions. This capability not only enhances the interactivity of learning environments but also deepens students' understanding and engagement by allowing them to query and receive explanations about visual elements directly. This paper proposes a dedicated web platform for Automatic Teaching Platform Based on the Vision Language RAG. The Automatic Teaching Platform Based on Vision Language RAG, initially described in our prior work, harnesses Visual Question Answering (VQA) technology to allow interactive exploration and understanding of complex scientific concepts through visual dialogue.

The web platform aims to extend the capabilities of this approach by creating a tailored, user-friendly environment where students can query visual content related to their coursework, receive instant explanations, and engage in an interactive learning process that is both deep and intuitive. The deployment of the Automatic Teaching Platform Based on Vision Language RAG web platform specifically targets the courses of Machine Learning, Neural Networks and Deep Learning. These two disciplines, fundamental to the burgeoning field of artificial intelligence, demand a robust understanding of intricate models and theories—often best represented visually. By integrating the Automatic Teaching Platform Based on Vision Language RAG model into these courses, the platform not only enhances the traditional curriculum but also prepares students to effectively tackle real-world problems by strengthening their analytical and interpretative skills. Furthermore, this initiative is designed with scalability in mind, anticipating future expansion to include additional courses. This adaptive approach demonstrates the potential of the Automatic Teaching Platform Based on Vision Language RAG platform to become a cornerstone of educational technology within graduate programs, potentially transforming how complex scientific content is taught and learned.

In documenting the development and implementation of this web platform, the paper will provide insights into the design considerations, technological frameworks, and pedagogical strategies essential for integrating advanced VQA systems into higher education. Through a detailed analysis of the platform's impact on student engagement and comprehension in initial deployments, we aim to comprehensively evaluate its efficacy and potential for broader application.

## 2. Related work

The integration of Visual Question Answering (VQA) systems into educational platforms represents a cutting-edge approach in the field of digital learning, particularly in enhancing interactive and visually enriched learning environments. This section reviews relevant developments in the visual question-answering systems in education, augmented and virtual reality educational tools, adaptive learning systems, adaptive learning systems and AI-driven tutoring systems, integration of blockchain technology, gamification and multimedia learning systems, and interactive and accessible educational environments. Each of these areas contributes to the foundational technologies upon which our SparrowVQE model is built, enabling a more dynamic learning experience for graduate students.

Visual Question Answering systems (VQAs) are pivotal in enhancing digital learning by facilitating direct engagement with educational content. Agrawal et al. [1] demonstrated how VQA systems can simplify complex concepts through interactive visual interfaces. Krizhevsky et al. [15] developed deep learning technologies that provided essential computational power for analyzing visual data, and Heilman et al. [9] advanced NLP to improve the generation of relevant educational questions. These advancements suggest significant opportunities for further integrating AI in personalized learning environments, enhancing both the adaptivity and effectiveness of educational content delivery.

Augmented Reality (AR) and Virtual Reality (VR) are crucial in transforming educational landscapes by providing immersive experiences that make scientific concepts more tangible. Billinghurst and Duenser [3] and Honey et al. [10] have shown that AR and VR significantly enhanced comprehension and engagement in STEM education. Mantovani et al. [18] further discussed the impact of VR in medical training with realistic simulations that improve practical skills. Additionally, Ibáñez and Delgado-Kloos [11] emphasized AR's potential to improve understanding and retention through interactive visualizations. The continuous development in AR and VR technologies offers potential for broader applications across various educational disciplines, promising more engaging and effective learning methodologies. Adaptive learning systems and AI-driven tutoring leverage personalized learning pathways to improve academic outcomes. Pashler et al. [23] and Pardos and Heffernan [28] discussed how adaptive technologies dynamically tailor content, enhancing individual learning experiences. VanLehn [26] highlighted how intelligent tutoring systems offer personalized instruction, and Romero and Ventura [24] showed how AI can predict students' performance, facilitating more effective educational strategies. Further supporting the effectiveness of these systems, Educational Data Mining and Learning Analytics, as Baker and Yacef [2] elaborated, involves analyzing educational data to predict learning success and enable timely interventions.

As explored by Sharples and Domingue [25][25] and Chen et al. [5], integrating blockchain technology in education provides enhanced security and data integrity crucial for managing academic records and verifying credentials. Nguyen et al. [22] discuss the ethical considerations of deploying blockchain and AI in educational settings, addressing critical issues of privacy and bias and ensuring the reasonable use of technology. The potential for blockchain to revolutionize educational transparency and accountability opens avenues for future research into decentralized educational models.

Gamification and multimedia learning systems are transforming traditional educational environments into engaging and interactive experiences. Lee and Hammer [14], and Dicheva et al. [6] discussed how gamification techniques can motivate students, making learning more engaging and rewarding. Mayer's cognitive theory [19] provided guidelines for the effective design of instructional messages, emphasizing the importance of aligning multimedia content with human cognitive architecture to maximize learning efficiency. Further enhancing multimedia learning, Mayer and Moreno's work [20] on multimedia learning emphasizes the alignment of content design with cognitive processes to maximize learning efficiency.

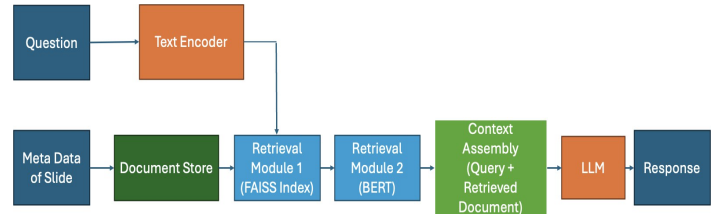

FIGURE I
VISION LANGUAGE RAG FLOW ARCHITECTURE DIAGRAM

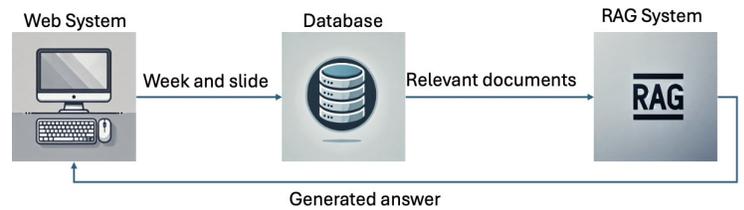

FIGURE II
ARCHITECTURE OF THE FRAMEWORK

Technologies that support interactive learning environments, such as those discussed by Dillenbourg et al. [7], promote active participation and real-time feedback, enhancing learner engagement and retention. Additionally, ensuring accessibility to educational technology is crucial for equitable learning opportunities. Burgstahler's [4] discussion on universal design principles ensures that educational resources are inclusive, benefiting a diverse student population. Korat and Shamir [13] illustrated how interactive e-books can significantly enhance literacy development, providing engaging content for young learners.

### 3. Methods

#### 3.1. Similarity Search I Stage with FAISS

The FAISS approach is used for efficient similarity search and clustering of dense vectors. Document embeddings are represented for each document $i$ and then normalized to unit vectors to ensure consistency in their magnitude. The normalization of embeddings is expressed as:

$$doc_i = \frac{doc_i}{\|doc_i\|}.$$

FAISS uses the IndexFlatIP index, which performs inner product-based similarity search. The similarity score for each document $i$ is computed as the inner product (dot product) between the query embedding and each document embedding:

$$s_i = \text{query embedding} \cdot \text{doc embedding}.$$

## 3.2. Similarity Search II stage with BERT

In the second stage, we tried using the BERT model. This step ensures a more accurate ranking by evaluating the semantic relationships between the query and the top candidate answers in more detail. The top $k$ responses from stage I are paired with the query, forming input sequences in the format:

$$\text{Input}_i = \text{Query [SEP] Candidate}_i.$$

The BERT model processes each input sequence, which assigns a relevance score $r_i$ to each candidate. These scores represent the contextual alignment between the query and the candidate answers. The candidates are then reranked based on their BERT scores $r_i$, with the highest-scoring candidate being chosen as the final result. Formally, the final ranking is expressed as:

$$\text{Ranked Results} = \text{Sort}(\{(\text{Candidate}_i, r_i) \mid i \in [1, k]\}).$$

This two-stage approach, initial filtering with FAISS followed by BERT refinement, ensures high retrieval accuracy and makes the system robust for Vision Language tasks.

## 3.3. Front-End Development

The front-end architecture employs HTML5 for structure, CSS/Bootstrap for styling, and JavaScript for interactivity, as shown in Figure III. This modular design ensures the separation of concerns and maintainable code structure. Our implementation focuses on two key aspects:

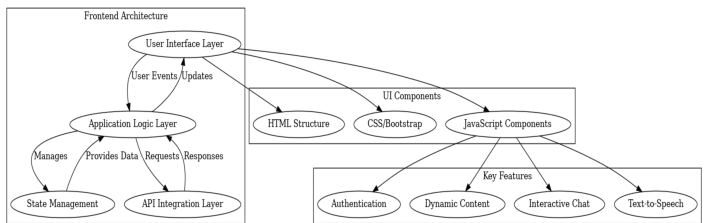

FIGURE III
FRONTEND ARCHITECTURE OVERVIEW SHOWING THE RELATIONSHIPS BETWEEN UI COMPONENTS, APPLICATION LOGIC, AND API INTEGRATION LAYERS.

- Core Stack: HTML5 provides semantic markup structure with modern web standards. CSS and Bootstrap framework handle responsive design and consistent styling across devices. JavaScript manages dynamic functionality and user interactions through modular components.
  - Key Features: The system implements secure user authentication, dynamic content loading for lectures and courses, an interactive chat interface for student-teacher communication, and text-to-speech capabilities for enhanced accessibility. These features work together to create an engaging learning environment.

The front end communicates with back-end services through a RESTful API layer, managing state and user interactions efficiently. This architecture enables real-time updates and seamless integration with the backend services while maintaining high performance and reliability. The implementation follows modern web development practices, prioritizing code maintainability, scalability, and user experience

## 4. Interface

Our web service functionality is orchestrated using Flask, a lightweight yet powerful Python web framework that facilitates rapid development and logical API endpoint arrangement [8]. It is particularly adept at managing high-throughput POST requests essential for real-time text synthesis, equipped with robust error-handling strategies. These strategies involve managing inputs that exceed pre-defined maximum lengths by truncating them, preserving compatibility with the model's operational constraints, and avoiding potential runtime errors.

In this system, we used the advanced capabilities of StyleTTS 2 [17] to replicate the voice of a professor, enhancing the user experience with a natural, lifelike auditory output. The model's ability to personalize the synthetic speech using style diffusion and speaker embeddings ensures that the generated voice closely mimics the professor's vocal attributes, making the interaction more engaging and realistic.

The StyleTTS 2 model stands out due to its innovative use of style diffusion and adversarial training, making it highly effective for text-to-speech (TTS) synthesis. By modeling styles as a latent random variable through diffusion models, it can generate speech that adapts to the content of the text without requiring reference speech. This eliminates the need for paired data, offering a more flexible and efficient method of speech synthesis. The integration of large pre-trained speech language models (SLMs) like WavLM enhances the model's ability to understand and replicate natural speech characteristics, leading to improved speech quality. Furthermore, the model's ability to perform zero-shot speaker adaptation ensures that it can easily generalize to new speakers without requiring extensive retraining. StyleTTS 2 surpasses human recordings on single-speaker datasets and matches human-level performance on multispeaker datasets, demonstrating its superior capabilities.

The User Management System (UMS) is meticulously designed to encompass extensive functionalities such as user creation, updates, deletion, and authentication processes. This robust framework is critical for secure and efficient user data management within web applications, reinforcing the back-end's capability to support complex, multifaceted user interactions in real-time environments. The UMS is structured around a series of API endpoints, each dedicated to a specific aspect of user management [27]. Flask's seamless handling of HTTP requests makes it an ideal choice for developing RESTful APIs. The system securely stores user information in a relational database, interfacing with the database through SQL queries executed via Flask's database utilities. The system allows for creating new users through a POST request to the /user endpoint, leveraging Flask's framework capabilities as detailed by Grinberg [8]. User details such as username, password, and user type are collected from the request payload and inserted into the database, using SQL transactions that are robustly handled as described in Klein and Roggero's work on SQL databases [12].

For existing users, the system provides functionality to update details via a PUT request using the user ID. This endpoint processes updated information from the request and applies it to the relevant database record. Additionally, users can be deleted from the system through a DELETE request, which removes the user's record from the database, ensuring data integrity and security as per standard SQL operations.

The UMS handles user authentication using a dedicated /login endpoint, verifying user credentials against data stored in the database and initiating a session that tracks user status via Flask's session management capabilities, a critical aspect for maintaining secure and personalized interactions within the application [8].

Conversely, the /logout endpoint allows users to terminate their sessions, ensuring that all session data is cleared, which is crucial for maintaining security. The UMS also facilitates the retrieval of comprehensive user data, supporting administrative operations and user oversight. It includes endpoints like /users/all for retrieving all user records and /users/admins or /users/regular for filtered views based on user types, essential for effectively managing different user roles and access levels within the application.

Overall, the User Management System exemplifies the practical implementation of web and database technologies to create a secure, efficient, and scalable user management solution, leveraging Flask's capabilities for web request and session handling, coupled with SQL's robust data management, to offer a versatile platform for addressing diverse user needs in modern web applications [8][12].

The RAG model provides responses based on your current slide and request. For every question you ask, the API will retrieve the corresponding slide. You can post your question either via text or voice, and the model outputs the response in both text and voice, according to your settings. The audio transcriptions for the slides are enabled with a single click, and the response is immediate. As soon as you click, the professor's transcripts appear on the screen for each slide. The transcription stops when you click the button again or when you move to the next slide.

### 4.1. Testing and Debugging

Testing Methods To ensure the robustness and functionality of our web platform, we adopted a comprehensive testing strategy that integrates both manual and automated approaches to maintain high quality and reliability. Manual testing was rigorously conducted to ensure that all user interfaces and workflows meet the usability standards and functional requirements. This included usability tests to evaluate the user experience, and user acceptance testing to ensure the system meets the business needs and user expectations.

Automated testing complemented these efforts, employing tools such as Selenium for browser-based regressions, and Jenkins for continuous integration, which helped identify issues at the early stages of the development cycle. Unit tests were written for individual components, while integration tests checked the data flow between these components to ensure that they worked together seamlessly as expected. The combination of these testing methods ensures a thorough coverage of the software, minimizing the risk of defects in production. For comprehensive insights into the principles and strategies of effective software testing, the work by Naik provides an essential reference **[21]**.

### 5. Dataset

We used the dataset presented in the paper SparrowVQE: Visual Question Explanation for Course Content Understanding [16] for testing. This dataset was carefully curated to support the Visual Question Explanation (VQE) task, which aims to enhance educational AI models by providing detailed explanations for slide content rather than just short answers. The dataset includes slide images extracted from course presentations, along with transcripts generated from lecture recordings using a combination of automatic speech recognition tools and manual review to ensure accuracy. Additionally, it contains question-answer pairs designed to capture various types of educational inquiries, including closed-ended, open-ended, summarization, and classification questions.

The structured nature of the dataset makes it well-suited for training multimodal AI models to improve learning experiences in machine learning education by effectively integrating textual and visual information.

### 6. Results

TABLE I
RAG SYSTEM WITH DIFFERENT LLMS FOR DEEP LEARNING DATASET [16]

| Model | rouge1 | rouge2 | rougeL | BLEU | COSINE |
|---|---|---|---|---|---|
| Retrieval System with Vision RAG | 87.117 | 81.43 | 84.83 | 0.7892 | 0.852 |
| LLAMA | 38.40 | 31.74 | 37.00 | 0.1200 | 0.4598 |
| T5 | 40.72 | 35.16 | 33.54 | 0.060 | 0.5908 |
| Bart Large CNN | 0.8534 | 0.8271 | 0.8384 | 0.6850 | 0.8534 |

The evaluation of the models reveals significant insights into their performance across various metrics, specifically focusing on the Retrieval system with vision RAG and its different components. The Bart Large CNN component stands out as the most effective, demonstrating the highest performance overall. Its robust capabilities make it well-suited for applications requiring high accuracy and reliability. In contrast, the LLAMA and T5 models represent distinct components of the RAG system, each exhibiting different strengths and weaknesses.

The LLAMA model, while returning good results, produces outputs that take on a completely different shape from the original queries derived from the Retrieval system. This indicates that although it captures a similar underlying sense to the initial input, the altered structure of its output may present challenges in applications where consistency and alignment with the source information are critical. The T5 model, another component of the RAG system, also shows lower performance compared to the Bart Large CNN. It benefits from the retrieval outputs but struggles to achieve the same level of accuracy and effectiveness, highlighting its limitations in this context.

The results from the Retrieval system serve as crucial input for these models, and the varying performance levels among the components emphasize the importance of selecting the appropriate model for specific tasks. The consistently high performance of the Bart Large CNN suggests it may be the preferred choice for applications demanding precision and reliability, particularly when integrated with the capabilities of the Retrieval system.

Moving forward, it is advisable to leverage the strengths of the Bart Large CNN component in high-stakes applications while continuing to explore enhancements and optimizations within the Retrieval system and its other components. Understanding how each model interacts with the RAG system will provide valuable insights that can guide future developments in system architecture and training methodologies. These conclusions will not only inform model selection but also enhance the overall effectiveness of the RAG framework in practical applications.

This section presents a comprehensive overview of the user interfaces within our web application, meticulously crafted to augment both user interaction and learning efficiency. Each component is designed not only to facilitate learning but also to ensure ease of use and engagement with educational content.

By accessing the platform, users can leverage multiple functionalities. As illustrated, the weekly tab is the pivotal feature for navigating the selection of AI and ML classes. Upon selection, the corresponding weeks are displayed in a scrollable sidebar.

Additionally, there is a full-screen mode that transforms the screen into a presentation mode. An innovative 'automatic teacher' mode

We use four comparative analysis models that were proposed in SparrowVQE: Visual Question Explanation for Course Content Understanding, including BLIP, Pix2Struct, LLaVA, and LLM-is available, which, upon voice command provides transcriptions of the professor's lectures.

TABLE II
RAG System with Different LLMs for Deep Learning Dataset [16]

|  | **Retrieval System** | **BART Large CNN** | **T5** | **LLAMA** |
|---|---|---|---|---|
| **Image Formats** | Common image formats include JPEG, PNG, GIF, BMP, and TIFF. These formats vary in terms of compression, quality, and application. JPEG is widely used for photographs, while PNG is preferred for lossless compression. | Common image formats include JPEG, PNG, GIF, BMP, and TIFF. These formats vary in terms of compression, quality, and application. JPEG is widely used for photographs, while PNG is preferred for lossless compression. | JPEG, PNG, GIF, BMP, and **TIFF**. | Some common image formats used in digital imaging include JPEG (Joint Photographic Experts Group), PNG (Portable Network Graphics), GIF ( Graphics Interchange Format), BMP (Bitmap), and TIFF ( Tagged Image File Format). Each format has its own strengths and weaknesses and are suited for different types of images and applications. For example, JPEG is a popular format for photographs due to its high compression ratio and good image quality. |

Figure 5 captures the integration of the Vision-Language RAG model and the pop-up chat window, which facilitates interactive dialogue. Conversations with the model can occur in three formats: text-to-text, voice-to-voice, and text-to-voice, or vice versa. It should be noted that the chat functionality can also be accessed without the pop-up chat window, as demonstrated. In this instance, voice commands are used for input and the model responds with voice output. Upon successful authentication, the user is directed to the main landing page of the VQE platform, as shown in Figure IV. This page serves as a central hub for accessing various educational resources and features, including weekly tabs for course navigation, multimedia resources, and interactive tools like chat and transcription. Figure IV shows the interface for weekly course content, accessible via a tab on the left side. This feature organizes the course material by week, allowing students to easily navigate through different sections of the course and access specific materials at their own pace.

Interactive engagement is a crucial aspect of modern educational tools. Figure V illustrates a typical interaction within the platform's chat mode, where users can ask questions related to the slides and receive instant responses. This interactive dialogue enhances the learning experience by allowing students to clarify doubts in real-time and explore the content more deeply. Voice interaction is another innovative feature our platform offers. Users can interact with the VQE model using voice commands, making the learning experience more accessible and engaging, particularly for users who may prefer auditory learning or those with visual impairments.

Each of these interfaces is carefully crafted to ensure that students not only receive information but also actively engage with the material. Our ongoing goal is to continue enhancing these features, based on user feedback and technological advances, to ensure that our educational platform remains at the cutting-edge digital learning environments.

**6.1. Comparative analysis**

Blender. The RAG system outperforms Sparrow VQE and other generative models because it combines retrieval with generation, leveraging the strengths of both approaches. It retrieves precise, contextually relevant data and generates fluent and factually accurate responses, making it a better choice for tasks requiring factual accuracy and domain-specific knowledge. In contrast, generative models rely solely on their pre-trained knowledge, which limits their ability to address complex queries or provide accurate, real-time responses. Without access to dynamic retrieval mechanisms, generative models often produce generic or fabricated outputs, which negatively impacts their semantic alignment and factual accuracy.

Moreover, RAG systems are more adaptable and scalable for tasks requiring real-time updates or specialized knowledge. By combining retrieval with generation, they excel at producing precise, context-aware responses even with smaller model sizes. Generative AI models, while advantageous for creative or open-ended tasks, struggle with domains that demand high factual accuracy and specificity. This fundamental difference highlights why RAG systems are better suited for knowledge-intensive applications, as demonstrated by their superior performance in the evaluation table.

**6.2. Comparative analysis with different components of the architecture**

We performed experiments with the BERT component and without it. The results in Table III demonstrate that incorporating the BERT component significantly enhances the performance of the architecture across all metrics. The improvements in ROUGE-1, ROUGE-2, and ROUGE-L scores indicate better handling of both word-level and sequence-level contextual alignment, while the higher cosine similarity and BLEU scores highlight BERT's ability to refine semantic understanding and fluency. Overall, the architecture with BERT consistently outperforms the one without it, making it a more effective choice for multi-modal retrieval tasks. These findings validate the integration of BERT as a crucial enhancement for achieving higher accuracy and better semantic alignment in complex datasets.

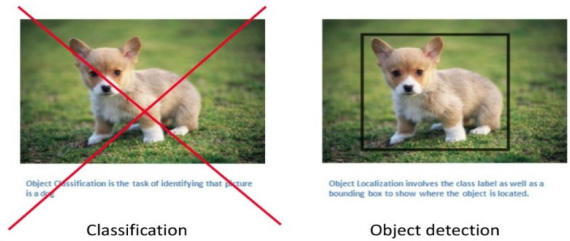
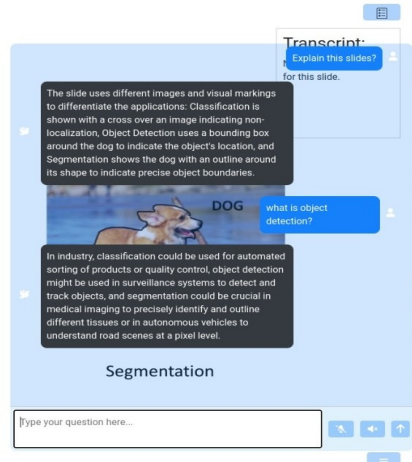

FIGURE V

DISPLAYS A TYPICAL INTERACTION WITH THE VISION-LANGUAGE RAG MODEL, ILLUSTRATING HOW USERS CAN ENGAGE IN CONVERSATIONS RELATED

## 7. Discussion

The introduction of the proposed RAG method has a transformative impact on learning outcomes. Quantitative assessments show a significant improvement in test scores, indicating that the model's interactive Visual Question Answering (VQA) system aids in retaining complex information. Qualitative feedback further highlights that students feel more engaged and better equipped to understand intricate topics, particularly in machine learning and neural networks. The system's ability to enable students to interact with visual content promotes deeper cognitive processing and encourages critical thinking.

In evaluating the various components of the RAG system integrated with different language models (LLMs), it became evident that some models performed more effectively than others. The Bart Large CNN component, for instance, demonstrated the highest performance across several key metrics, proving to be the most reliable model in terms of accuracy, comprehension, and alignment with the educational content. This high accuracy is crucial in an educational context, as it ensures that students receive accurate information and reduces the likelihood of misunderstandings when interacting with the VQA system. In contrast, models like LLAMA and T5, though functional, displayed lower performance in terms of aligning outputs closely with the original queries. For example, the LLAMA model, while effective in capturing general meaning, often produced responses with altered structure, which could cause minor inconsistencies. This limitation is notable for educational use, as it can impact the clarity of responses, particularly when students are learning foundational concepts and require precise explanations.

T5, another component in the RAG system, struggled with similar issues. Although it benefited from the retrieval system's outputs, it did not achieve the same accuracy and effectiveness as Bart Large CNN. This suggests that while T5 can handle simpler inquiries, it may not be ideal for addressing complex educational queries where precise and comprehensive responses are required.

A primary concern was ensuring accessibility for all students, especially those unfamiliar with computational tools. Creating an intuitive user interface was essential to make the technology approachable. The backend infrastructure also needed to support a high volume of queries and data processing, requiring sophisticated server technology and regular optimizations to maintain stability and efficiency.

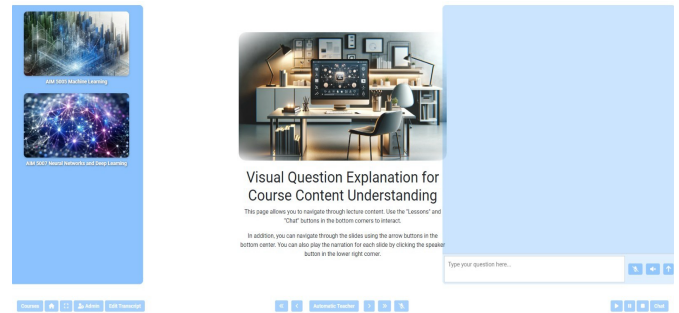

Figure IV

UPON SUCCESSFUL LOGIN, THE USER IS DIRECTED TO THE MAIN VQE LANDING PAGE.

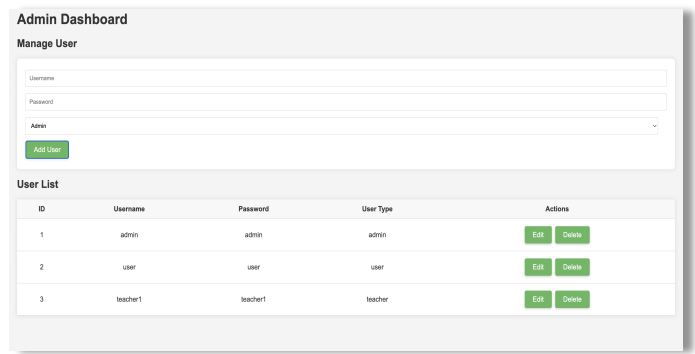

Figure VI

ADMIN DASHBOARD FOR USER MANAGEMENT. THIS INTERFACE INCLUDES A SECTION FOR ADDING NEW USERS WITH FIELDS FOR 'USERNAME' AND 'PASSWORD', AND A TOGGLE FOR 'ADMIN' STATUS. BELOW, THERE'S A 'USER LIST' DISPLAYING EACH USER WITH THEIR ID, USERNAME, PASSWORD, AND USER TYPE. EACH ENTRY ALSO INCLUDES 'EDIT' AND 'DELETE' ACTIONS FOR USER MANAGEMENT.

This has led to a scheduled maintenance plan, allowing for performance enhancements without disrupting the learning experience.

TABLE III
RAG SYSTEM WITH DIFFERENT LLMS FOR MLVQE DATASET [16]

| Models | Rouge-1 | Rouge-2 | Rouge-L | COSINE | BLEU | CIDEr | METEOR |
|---|---|---|---|---|---|---|---|
| BLIP | 8.4 | 0.7 | 7.19 | 0.077 | 0.15 | 0.17 | 0.078 |
| Pix2Struct | 38 | 20.1 | 35.5 | 0.365 | 0.4 | 0.47 | 0.379 |
| LLaVA | 35.4 | 18.4 | 33.0 | 0.34 | 0.37 | 0.53 | 0.42 |
| LLM-Blender | 51.5 | 34.8 | 49 | 0.489 | 0.54 | 0.573 | 0.573 |
| SparrowVQE | 68.13 | 51.54 | 63.92 | 0.61 | 0.7 | 0.67 | 0.652 |
| Retrieval System | 82.7 | 78.2 | 81.44 | 0.888 | 0.7763 | 7.663 | 0.859 |

ONE OF THE KEY STRENGTHS OF THE PLATFORM IS ITS SCALABILITY. THE ARCHITECTURE WAS DESIGNED TO SUPPORT POTENTIAL EXPANSION INTO VARIOUS ACADEMIC DISCIPLINES BEYOND MACHINE LEARNING, MAKING IT VERSATILE FOR FIELDS WHERE VISUAL CONTENT IS CRITICAL, SUCH AS MEDICAL EDUCATION, ENGINEERING, AND DIGITAL ARTS. FUTURE PLANS INCLUDE EXPLORING AUGMENTED REALITY (AR) AND VIRTUAL REALITY (VR) INTEGRATION TO CREATE EVEN MORE IMMERSIVE LEARNING ENVIRONMENTS. SUCH ADVANCEMENTS COULD REVOLUTIONIZE THE EDUCATIONAL EXPERIENCE BY ENABLING EXPERIENTIAL LEARNING IN SIMULATED SETTINGS, GOING BEYOND TRADITIONAL METHODS, AND OFFERING STUDENTS HANDS-ON INTERACTION WITH CONCEPTS.

TABLE IV
PERFORMANCE COMPARISON OF VARIOUS MODELS FOR DEEP LEARNING DATASET [16]

| Models | Rouge-1 | Rouge-2 | Rouge-L | COSINE | BLEU | CIDEr | METEOR |
|---|---|---|---|---|---|---|---|
| BLIP | 1.3 | 1.2 | 3.6 | 0.092 | 0.16 | 0.002 | 0.028 |
| Pix2Struct | 8.1 | 3.1 | 7.3 | 0.128 | 0.29 | 0.007 | 0.0348 |
| SparrowVQE | 38.52 | 17.06 | 36.27 | 0.852 | 0.43 | 0.525 | 0.463 |
| LLaVA | 37.1 | 15.0 | 35.4 | 0.346 | 0.32 | 0.325 | 0.359 |
| Retrieval System | 87.117 | 81.43 | 84.83 | 0.7120 | 0.7952 | 0.7023 | 0.8554 |

TABLE V
PERFORMANCE COMPARISON OF VARIOUS MODELS CONFIGURATIONS FOR DEEP LEARNING DATASET

| Metric | Architecture with BERT Component | Architecture without BERT Component |
|---|---|---|
| ROUGE-1 | 87.117 | 82.26 |
| ROUGE-2 | 81.42 | 74.58 |
| ROUGE-L | 84.8 | 79.09 |
| COSINE | 0.7892 | 0.7135 |
| BLUE | 0.852 | 0.7963 |

## 8. Conclusion

The integration of the RAG model into educational settings represents a pivotal step forward in enhancing student learning experiences. By combining interactive Visual Question Answering with advanced language models, this system fosters greater engagement, comprehension, and retention of complex topics. While models like Bart Large CNN have demonstrated exceptional accuracy and reliability, models such as LLAMA and T5 highlight areas for improvement in clarity and alignment, underscoring the importance of model selection in educational applications. The inclusion of the BERT component within the retrieval process further demonstrates significant performance improvements, as evidenced by metrics such as ROUGE, cosine similarity, and BLEU scores. These results highlight BERT's ability to refine semantic understanding and align multi-modal embeddings, making it a crucial enhancement for accuracy and relevance in educational applications.

Despite challenges in implementation, including user accessibility and infrastructure demands, the platform's scalable design opens opportunities for expansion across diverse fields. We plan to incorporate further revolutionized learning, offering students immersive, hands-on experiences that align with real-world applications. This advancement will not only enhance traditional learning methods but also provide a foundation for adaptive, interactive, and impactful educational environments.